\documentclass[runningheads]{llncs}
\usepackage{graphicx}
\usepackage{comment}
\usepackage{amsmath,amssymb} % define this before the line numbering.
\usepackage{color}
\usepackage{subcaption}
\begin{document}
% \linenumbers
\pagestyle{headings}
\mainmatter

\title{Info3D: Representation Learning on 3D Objects using Mutual Information Maximization and Contrastive Learning} % Replace with 

%******************

%\begin{comment}
\titlerunning{Info3D}
\author{Aditya Sanghi }
\authorrunning{A. Sanghi }
\institute{Autodesk AI Lab, Toronto, Canada \\ \email{aditya.sanghi@autodesk.com}
}

%\end{comment}
%******************
\maketitle

\begin{abstract}
A major endeavor of computer vision is to represent, understand and extract structure from 3D data. Towards this goal,  unsupervised learning is a powerful and necessary tool. Most current unsupervised methods for 3D shape analysis use datasets that are aligned, require objects to be reconstructed and suffer from deteriorated performance on downstream tasks. To solve these issues, we propose to extend the InfoMax and contrastive learning principles on 3D shapes. We show that we can maximize the mutual information between 3D objects and their ``chunks" to improve the representations in aligned datasets. Furthermore, we can achieve rotation invariance in  SO${(3)}$ group by maximizing the mutual information between the 3D objects and their geometric transformed versions. Finally, we conduct several experiments such as clustering, transfer learning, shape retrieval, and achieve state of art results.

\end{abstract}

\section{Introduction}

\begin{figure}[t]
\begin{center}

 \includegraphics[width=0.8\linewidth]{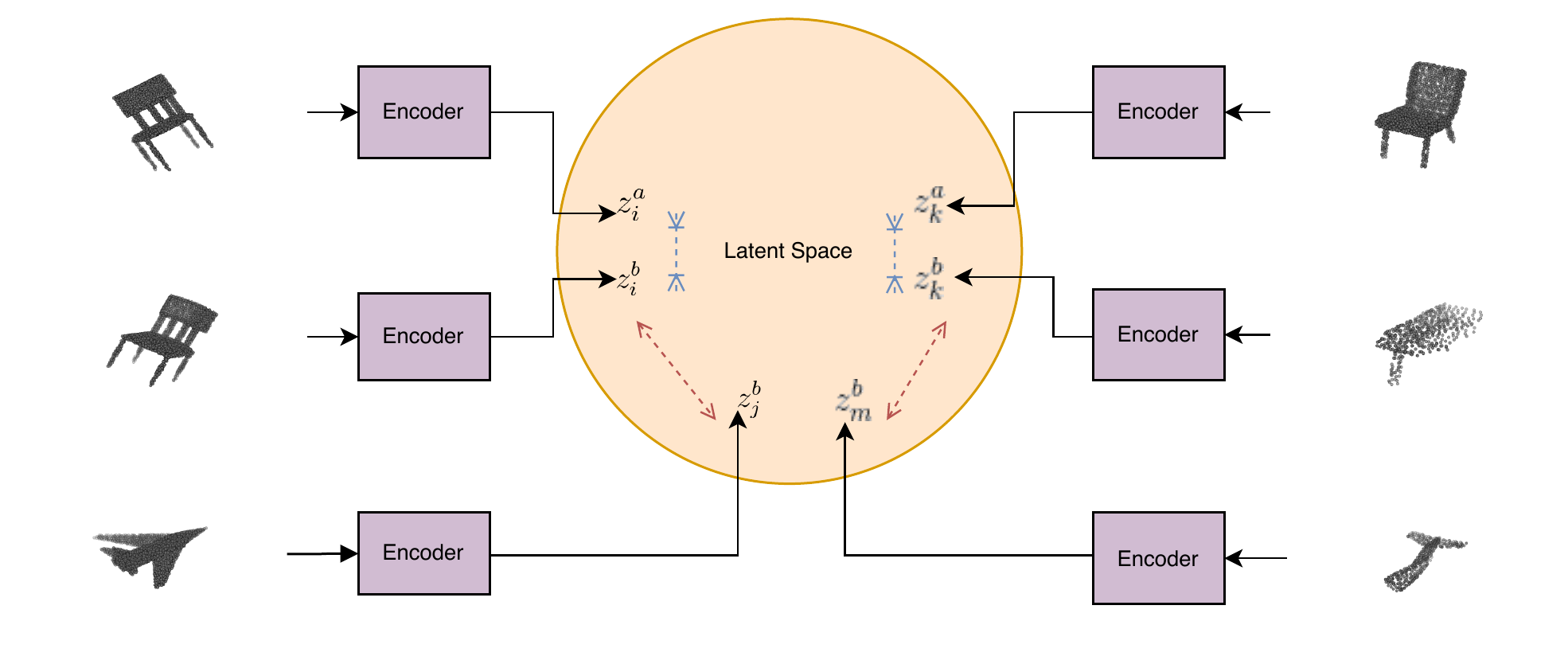}
\end{center}
   \caption{\textbf{General idea of the method.} We try to bring the 3D shape and the different view of the 3D shape closer in latent space while pushing away the representation of other objects in the dataset further away.  }
\label{fig:long}
\label{fig:fig1}
\end{figure}

Recently, several unsupervised methods have managed to extract powerful features for 3D objects such as in  \cite{groueix2018atlasnet}, \cite{yang2018foldingnet}, \cite{achlioptas2017learning}, \cite{park2019deepsdf} and \cite{mescheder2019occupancy}. However, these methods assume all 3D objects are aligned and have the same pose in the given category. In real world scenarios, this is not the case. For example, when a robot is identifying and picking up an object, the object is in an unknown pose. Even in online  repositories of 3D shapes, most of the data is randomly oriented as users create objects in different poses.  To use these methods effectively we require to align all objects for a given category which is a very expensive and time consuming process.  

Furthermore, these unsupervised methods require reconstruction of 3D shapes which is not ideal for many reasons.  Firstly, it is not always feasible to reconstruct the 3D representation of a shape. For example, due to the discrete nature of meshes, reconstructing their representation may not be attainable. Moreover, in cases where we need invariant representation, it's hard to reconstruct back the shape from this invariant representation. For example, if you wanted to create rotation invariant embeddings, you need to lose pose information from the embeddings of 3D objects. If you lose the pose information, it will not be possible to reconstruct the shape back as you need to reconstruct it with that given pose.

To overcome the challenges mentioned above we propose  a decoder-free, unsupervised representation learning mechanism which is rotation insensitive. The method we introduce takes inspiration from the Contrastive Predictive Coding \cite{oord2018representation} and Deep InfoMax \cite{velivckovic2018deep} approach. These methods usually require a different ``view" of the object, which is used to maximize the mutual information with the object. This other view can be different modalities, data augmentation of the object, local substructure of the object, etc. 

We consider two different views of a given object in this work. First, we consider  maximizing the mutual information between a local chunk of a 3D object and the whole 3D object. The intuition for using this view is that the 3D shape is forced to learn about its local region as it has to distinguish it from other parts of different objects. This greatly enhances the representation learnt in aligned objects. Second, we consider maximizing the mutual information between 3D shape and a geometric transformed version of the 3D shape. The advantage of maximizing the mutual information in this scenario is that it can create global geometry invariant representations. This is very useful in the case of achieving rotation insensitive representation in SO$(3)$. Figure \ref{fig:fig1} illustrates the rough intuition behind the method. Note, despite using objects from different category in the figure, we push away every other shape in the dataset. This might even include object from the same category.  This method can be thought as instance discrimination \cite{wu2018unsupervised}.

The key contributions of our work are as follows:
\begin{itemize}
  \item We introduce a decoder-free representation learning method which can easily be extended to any 3D descriptor without needing to construct complex decoder architectures.  
  \item We show how local chunks of 3D objects can be used to get very effective representations for downstream tasks. 
  \item We demonstrate the effectiveness  of the method on rotated inputs and show how it is insensitive to such rotations in SO$(3)$ group.
  \item We conduct several experiments to show the efficacy of our method and achieve state of art results in transfer learning, clustering and semi-supervised learning. 
\end{itemize}

\section{Background}
\noindent
\textbf{Representation Learning on 3D objects.} Much progress has been made to learn good representations of 3D objects in an unsupervised manner which can then be used in several downstream tasks. In point clouds, works such as \cite{achlioptas2017learning}, \cite{yang2018foldingnet}, \cite{groueix2018atlasnet}, \cite{hassani2019unsupervised}, \cite{sauder2019context} and \cite{zhao20193d} have been proposed. For voxels and implicit representations, work such as \cite{wu2016learning}, \cite{mescheder2019occupancy}, \cite{chen2019learning}, \cite{michalkiewicz2019deep}, \cite{park2019deepsdf} create powerful representation features which are used for several downstream tasks such as shape completion, shape generation and classification. Recently, there has been a lot of progress in auto-encoding meshes such as in  \cite{tan2018variational}, \cite{cheng2019meshgan}, \cite{tan2018variational}. One disadvantage of  the above approaches are that they require you to reconstruct or generate the 3D shapes. As stated earlier, it might be expensive or not possible to reconstruct the shape. A recent approach \cite{zhang2019unsupervised} does not require to reconstruct the shape and instead does two stages of training. It first uses parts of a shape to learn features using contrastive learning and then uses pseudo-clusters to cluster all the data. Our method does not require two stages of training, and furthermore allows us to create rotation invariant embeddings. 
\linebreak

\noindent
\textbf{Maximizing Mutual Information and Contrastive Learning.} A lot of methods have used mutual information to do unsupervised learning. Historically, works such as \cite{linsker1989application},  \cite{becker1992information}, \cite{bell1995information}, \cite{becker1996mutual} have explored the InfoMax principle and mutual information maximization. More recently \cite{oord2018representation} proposed the Contrastive Predicting Coding framework which uses the embeddings to capture maximal information about future samples. The Deep InfoMax (DIM) \cite{velivckovic2018deep} approach is similar  but has the advantage of doing orderless autoregression. In concurrent to these works, works such as \cite{wu2018unsupervised} and \cite{zhuang2019local}, have extended these ideas from the metric learning point of view. These methods were extended in \cite{bachman2019learning}, \cite{henaff2019data},  \cite{tian2019contrastive} by considering multiple views of the data and achieved state of the art representation learning on images. The InfoMax principle has been extended to graphs \cite{velivckovic2018deep}, \cite{sun2019infograph} and for state representation in reinforcement learning \cite{anand2019unsupervised}. Our method is inspired by these methods and we extend them to 3D representations. For the multiple views, we use geometric transformations and chunks of a 3D object. Finally, in concurrent to our work several news works such as \cite{chen2020simple}, \cite{misra2019self} have been proposed. 
\linebreak

\noindent
\textbf{Rotation Invariance on 3D objects.} Traditionally, several methods have focused on hand engineered features to get local rotation invariant descriptors, such as \cite{rublee2011orb}, \cite{steder2010narf}, \cite{lazebnik2004semi}. More recently, methods such as \cite{deng2018ppf} and \cite{deng20193d}, first, get local invariant features by encoding local geometry into patches and then use autoencoder to reconstruct the local features. These approaches require creating hand-crafted local features and need normal information. Methods such as MVCNN \cite{su2015multi}, Rot-SO-Net \cite{li2019discrete} and \cite{sanghitowards} explicitly force  invariance by taking multiple poses of the object and aggregating them over the poses. However, such methods only work on discrete rotations or can only reconstruct objects rotated along one axis. A lot of deep learning methods have also attempted to use equivariance based architectures to achieve local and global rotation equivariance in 2D and 3D data. Methods such as \cite{cohen2016group}, \cite{worrall2017harmonic}, \cite{esteves2019equivariant}, \cite{thomas2018tensor} either use constrained filters that achieve rotation equivariance or use filter orbit which are themselves equivariant. It is usually difficult to create such architectures for different 3D representations. Furthermore, to generate invariant representation to rotation from equivariant representation usually requires a post-processing step. Our method uses the InfoMax loss with rotation transformation to enforce rotation invariance. This method can be easily extended to voxels, implicit representation and meshes without needing to create complex architectures.

%------------------------------------------------------------------------
\section{Methodology}
%-------------------------------------------------------------------------

\begin{figure*}
\begin{center}
\includegraphics[width=0.99\linewidth]{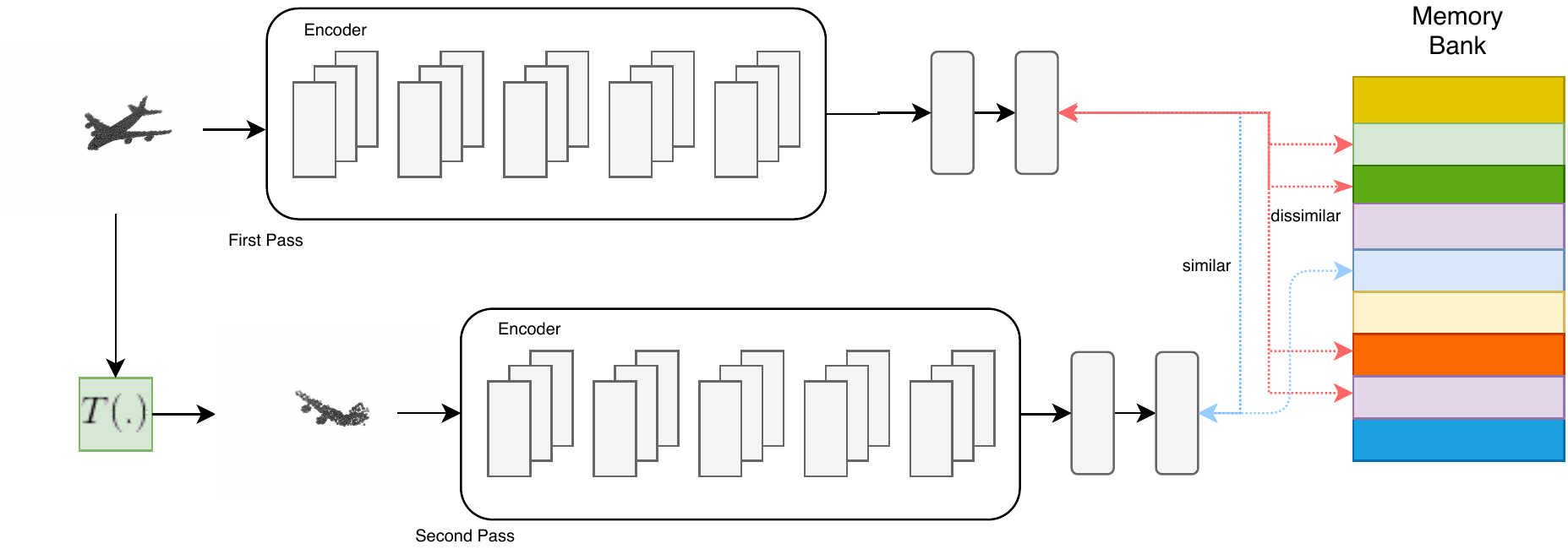}
\end{center}
   \caption{\textbf{Overview of the method.} A 3D object and different view of that object is encoded using the same encoder. The features across these views are made similar while a memory bank is used for negative examples. We also store the features obtained in the memory bank. }
\label{fig:fig2}
\end{figure*}

Our goal is to extract good features from 3D shapes in an unsupervised manner. To achieve this, we maximize the mutual information on the features extracted from a 3D shape and a different view of the 3D shape. We consider two such views in this work for achieving different purposes. First, we consider using a ``chunk" of a 3D object to improve the representation of aligned shapes. The goal of using chunks of a 3D object is to incorporate some form of locality and structure in input into the objective. We investigate different ways to extract a chunk of a 3D object and show how this can improve the representations learned. Next, we consider geometric transformation as the second view. Intuitively, the goal is to make the shape and its geometric transformed version closer in the latent space while distancing itself from other objects. We show how this can create transformation invariant embeddings to a large extent. We discuss this method in the context of point clouds but it should be straightforward to extend it to other 3D representations. The method in a pictorial form is shown in Figure \ref{fig:fig2}.

Let $x_i$ and $T(x_i)$ represent a 3D shape and the object obtained after applying the transformation as mentioned above. We use an encoder, $f(.)$, to transform $x_i$ and $T(x_i)$ into the latent representation $z_i^a$ and $z_i^b$.  Note that this encoder also includes a final normalizing layer which makes the embeddings unit vectors. In the next sections, we detail the motivation and mechanism for using chunks of a 3D object and geometric transformations as second view.

\subsection{Local Chunks}

\begin{figure*}
\begin{center}
\includegraphics[width=0.9\linewidth]{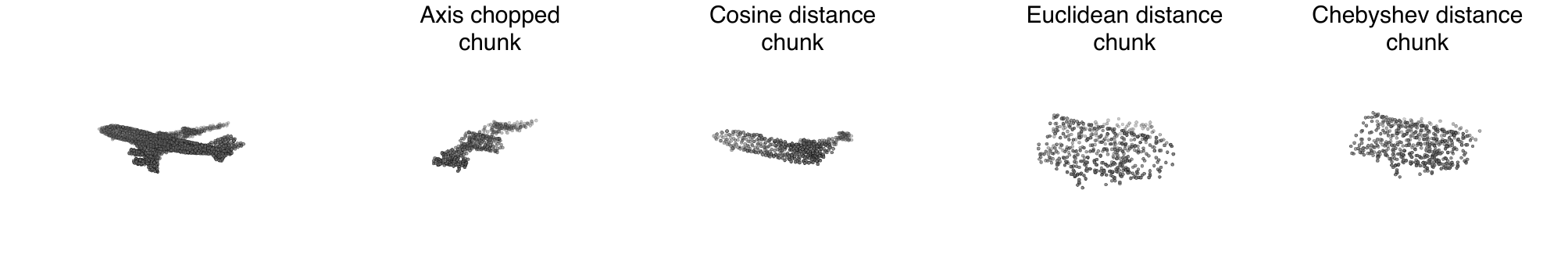}
\end{center}
   \caption{\textbf{Types of chunks.} The first object represents a sample 3D object. The rest of the objects are chunks obtained from that object. Note, for the last three chunks we use the same random point. Despite using the same random point, very different chunks can be obtained from different distance metrics. }
\label{fig:fig3}
\end{figure*}
As mentioned above, the chunks provide a way to define a locality structure to be incorporated in the objective. We do this by first defining a local subset of the 3D object. For point clouds, if you randomly select a subset of points, you will just get a coarser representation of the 3D object. So, we investigate some potential ways to define local sub-structures in point clouds. Next, we force the network to distinguish between its own local subset and other objects' local subsets using the InfoMax principle. This objective forces the network to learn about its own local sub-structures and creates more informative embeddings.       

We  define chunks by considering two mechanisms. In the first approach, we randomly select a point from the point cloud. Then, we use a distance measure in Euclidean space to select a subset of points. The distance measure we consider are Euclidean distance, cosine distance and Chebyshev distance. Once we select a subset of points, we normalize them using a bounding sphere. In the second approach, we take a chunk of a 3D object based on chopping the object randomly along the cartesian axes. The chunk is again normalized using a bounding sphere. The different chunks obtained from a sample 3D object are shown in Figure \ref{fig:fig3}.

\subsection{Geometric Transformation}
In this work, we consider several different geometric transformed versions of a 3D object. This can be thought of as a form of data augmentation. However, our method differs from traditional use of data augmentation by explicitly influencing the latent space to create better embeddings rather than implicitly hoping that it would create meaningful representation. Furthermore, we are doing data augmentation in an unsupervised setting, so the increased cost of augmentation can be shared across several  tasks instead of just one task such as in a supervised setting.  

 Though several other data augmentation methods can be used with the InfoMax principle, we consider geometric transformations for two major reasons. First, it is very trivial to apply an affine transformation to a 3D object. We simply need to multiply a transformation matrix.  Second, we can use the rotation affine transformation to create embeddings which are less sensitive to alignment. 
 
 In this paper, we only consider translation, rotation along $z$ axis, rotation in SO$(3)$ rotation group, uniform scale and non-uniform scale as our geometric augmentations. We also combine different transformations together to create more complex transformations. When we learn representations from unaligned datasets, we always rotate the object before applying a transformation to ensure it is not sensitive to rotation. 

\subsection{InfoNCE objective}

To estimate mutual information we use the InfoNCE \cite{oord2018representation} objective. Let us consider $N$ samples from some unknown joint distribution $p(x,y)$. For this objective, we need to construct positive samples and negative samples. The positive examples, are sampled from the joint distribution of $p(x, y)$ and negative samples from the product of marginals $p(x)$ $p(y)$. The objective is to learn a critic function, $h(.)$, by increasing the probability of positive examples and decreasing the probability for negative examples. The bound is given by 
\begin{equation}
I_{NCE} = \sum_{i=1}^{N}\log\dfrac{h(x_i, y_i)}{\sum_{j=1}^{N} h(x_i, y_j)}
\end{equation}

In our case, the positive samples are constructed by using the shape, $x_i$, and a different view, $T(x_i)$, of the shape. We construct negative examples by uniformly sampling, $k$ pairs, over the whole transformed version of the dataset. Note, this procedure can lead to objects from the same category being part of negative examples. We consider a batch size of $N$. The critic function is defined as exponential of bi-linear function of $f(x_i)$ and $f(T(x_i))$. We parameterize this function using $W$. Note that the critic can be defined on the global features from $f(.)$ or the intermediate features of $f(.)$. We can also modulate the distribution using the parameter $\tau$.  The critic function is shown below 
\begin{equation}
h(x_i , T(x_i)) = \exp(f(x_i)W f(T(x_i)/\tau)
\end{equation}

 We  now consider the objective where we maximize the mutual information between the global representations of $x$ and $T(x)$. That is, we maximize the mutual information between features from the latter layers of the encoder. The loss is shown below 
\begin{equation}
L = \sum_{i=1}^{N} -\log\dfrac{h(x_i , T(x_i))}{ h(x_i , T(x_i)) + \sum_{j=1}^{k}  h(x_i , T(x_j))}
\end{equation}

In theory, more negative examples, $k$, should lead to a tighter mutual information bound. One way of achieving this involves using large batch size, which might not be ideal. To avoid this, we take inspiration from \cite{wu2018unsupervised},  \cite{tian2019contrastive},  \cite{zhuang2019local} and use a memory bank to store data from previous batches. This allows us to use large number of negative examples. Increasing the number of negative examples leads to prohibitive cost in computing the softmax. Hence, we use the Noise-Contrastive estimation \cite{gutmann2010noise} to approximate the above loss as in \cite{tian2019contrastive}. The loss is as shown below

\begin{equation}
L_{NCE} =  \frac{1}{N} \sum_{i=1}^{N} \left(- \log{[h(x_i , T(x_i))] } - \sum_{j=1}^{k} \log{[1 -  h(x_i , T(x_j)]}\right)
\end{equation}

\section{Experiments}

We conduct several experiments to test the efficacy of our method and the representations learned by the encoder on both aligned and unaligned shape datasets. We divided the experiment section into three  parts. In the first section, we conduct experiments on aligned datasets and show the effectiveness of our method. In the second section, we discuss  representation learning on rotated 3D shapes. Finally, in the last section, we look at different hyperparameters and factors affecting our method.

\subsubsection{Training Details.}
For most of the experiments we use a batch size of 32, sample 2048 points on the shapes and use the ADAM optimizer \cite{kingma2014adam}. We use a learning rate of 0.0001. For ModelNet40, we run the experiment for 250 epochs in the case of aligned datasets whereas we run it for 750 epochs for unaligned datasets. As ModelNet10 is a smaller dataset, we run 750 epochs for aligned dataset and 1000 epochs for unaligned dataset. We set the number of negative examples to 512 and use 0.07 as temperature parameter. For ShapeNet v1/v2 dataset, we run it for 200 epochs and use 2000 negative examples. For aligned datasets we use the features from the 6th layer in our model whereas for unaligned datasets we use features from the 7th layer. Furthermore, for aligned datasets we use the cosine distance based chunks for clustering task and axis chopped chunks for all other experiments as the second view whereas for rotated datasets we use rotation in any SO$(3)$ rotation group plus translation as the second view. The choice of these parameters are further discussed in the ablation study section.  We also do early stopping if the network has converged. More details about the encoder structures and training details are in the appendix section. Finally, for ABC dataset \cite{koch2019abc} we use a batch size of 64 and use 1024 sample points on the surface. 

\subsubsection{Baseline setup.} For the tasks of clustering, rotation invariance and shape retrieval, we compare our method with three important works on representation learning on point clouds: FoldingNet \cite{yang2018foldingnet}, Latent-GAN Autoencoder  \cite{achlioptas2017learning} and AtlasNet  \cite{groueix2018atlasnet}. For all three baselines, we use similar training conditions as mentioned above, except we train the three models on ShapeNet \cite{chang2015shapenet} for 750 epochs and ModelNet40 \cite{wu20153d} for 1500 epochs for unaligned dataset. For aligned dataset we train ModelNet40 for 500 epochs. We use the PointNet encoder as the encoders for these baselines. Note, this is different from the respective paper implementations. More details regarding the architecture used for the baselines can be found in appendix. 

\subsubsection{Data Preparation.}
 All our experiments are conducted on ModelNet10 \cite{wu20153d}, ModelNet40 \cite{wu20153d}, ShapeNet v1/v2 \cite{chang2015shapenet} and ABC dataset \cite{koch2019abc}. In some of the above datasets objects are aligned according to their categories, so to unalign them we randomly generate a quanterion and rotate them in SO$(3)$ space. During unsupervised training, for each epoch,  we rotate the shape differently so that the network can see different poses of the same shape. However, when we test this on the downstream tasks, we only rotate the dataset once. This is to ensure that we only test the effectiveness of unsupervised learning part rather than the effectiveness of the downstream part.

\subsection{Representation Learning on Aligned Shapes}
In this section, we demonstrate how our method performs when we do representation learning on aligned shapes. Here we compare with well established baselines and show the advantages of our method. Moreover, we also show how the embeddings obtained from our method are more clusterable then autoencoder methods.  Note for clustering experiment we use the cosine distance based chunks whereas for all other experiments we use the axis chopped chunks as the second view.  

\subsubsection{Transfer Learning, semi-supervised learning and pre-training.}

\begin{table}
\begin{center}
\small
\begin{tabular}{|c|c|c|c|}
\hline
Unsup. Method & Acc. & Sup. Method  & Acc. \\
\hline\hline
3D-GAN \cite{wu2016learning} & 83.3  & PointNet \cite{qi2017pointnet} & 89.2 \\
Latent-GAN  \cite{achlioptas2017learning} & 85.7   & DeepSets \cite{zaheer2017deep} & 90.3 \\
ClusterNet \cite{zhang2019unsupervised}  &  86.8  & PointNet++ \cite{qi2017pointnet++} & 90.7 \\
FoldingNet \cite{yang2018foldingnet} & 88.4  & DGCNN \cite{wang2019dynamic} & 93.5 \\
Multi-Task PC  \cite{hassani2019unsupervised}  & 89.1  & Relational PC \cite{liu2019relation} & \textbf{93.6} \\
Recon. PC(PointNet) \cite{sauder2019context} &  87.3  &   PointNet (Pretrained) &  \textit{90.20} \\

Recon. PC(DGCNN) \cite{sauder2019context} & 90.6  &  DGCNN (Pretrained) &  \textit{93.03}  \\

Ours (PointNet)  & \textit{89.8}  &  & \\
Ours (DGCNN)  & \textbf{91.6}  &  & \\

\hline
\end{tabular}
\end{center}
    \caption{\textbf{Results on Aligned Datasets}. Left table represents the transfer learning results on ModelNet40 whereas the right table represents the supervised learning results.}
\label{tab:alignedresults}
\end{table}

\begin{table*}[t]
\begin{center}
\small
\begin{tabular}{|l|c|c|c|c|c|}
\hline
Method & 1 \%  & 2 \%  & 5 \% & 20 \%  &  100 \% \\
\hline\hline
FoldingNet \cite{yang2018foldingnet} &  56.15 & 67.05  & 75.97 & 84.06  &   88.41 \\
3D Cap. Net \cite{zhao20193d} &  59.24 &  67.67  &  76.49 & 84.48  &   88.91 \\

ours (best) & \textbf{59.66}  & \textbf{71.06}  & \textbf{80.48}  &  \textbf{87.66} & \textbf{91.64} \\
ours (mean) &  54.42 $\pm$ 3.77 & 66.34 $\pm$ 2.99  & 77.12 $\pm$ 1.51 &  86.851 $\pm$ 0.816  &  91.57 $\pm$ 0.06\\
\hline
\end{tabular}
\end{center}
\caption{\textbf{Semi-supervised results on ModelNet40.}}
\label{tab:semisuper}
\end{table*}

A well established benchmark used for unsupervised learning is transfer learning. We follow the same procedure as \cite{achlioptas2017learning} and \cite{yang2018foldingnet}. We first use unsupervised learning to train from the ShapeNet v1 dataset. We then train a Linear SVM using the training dataset of ModelNet40. In Table \ref{tab:alignedresults} we report the accuracy score on the test set of ModelNet40. Furthermore, we use the pre-trained weights from training ShapeNet dataset and initialize the pointnet classifier with those weights. We compare the results with randomly initialized weights by reporting the classification accuracy in Table \ref{tab:alignedresults}. Finally, we test our method in limited data scenarios. We compare our method to \cite{yang2018foldingnet} and \cite{zhao20193d} as mentioned in the appendix of \cite{zhao20193d}. It is not clear on how they select the subset of the data. This especially matters when we take very limited data, as you can have as less as 0 to 3 shapes per category. So we report both the best and mean accuracy over 10 runs of choosing a random subset. The results are reported in Table \ref{tab:semisuper}.   

It can be seen from the Table \ref{tab:alignedresults} that we  achieve state of the art results on transfer learning benchmark. We beat current state of the art by 1\% when we use DGCNN encoder with our method. Furthermore, using a simple encoder like PointNet beats many previous unsupervised methods with complex architectures and surprisingly beats the original PointNet supervised learning benchmark. Initializing our model with pre-trained weights also helps in achieving high accuracy in very less epochs. We can achieve 91\% accuracy within 3 epochs whereas it takes about 32 epochs for random initialize weights. This is shown in more detail in test accuracy training curve in the appendix. Finally, we perform very well in limited data scenarios. We can achieve about 87\% accuracy with 20\% of labelled data. 

\subsubsection{Clusterable representation.} A good unsupervised method would create a seperatable manifold associated with object classes \cite{bengio2013representation}. A good way to test this is to see how easily the data can be naturally clustered. We train the network by first using our unsupervised method and then use K-means algorithm on embeddings obtained. We use the implementation present in sklearn \cite{pedregosa2011scikit}. We set the number of clusters equal to 40 and use rest of the default parameters. To test the associations of the embeddings with the object classes we use adjusted  mutual information metric (AMI). The results are shown in the aligned column of Table \ref{tab:clusterALigned}. We use the training set of ModelNet40 for the embeddings.

\begin{table}
\begin{center}
\small
\begin{tabular}{|l|c|c|}
\hline
Method  & Aligned (AMI) & Unaligned (AMI) \\
\hline\hline
Latent-GAN  \cite{achlioptas2017learning}  & 0.646 $\pm$ 0.001  & 0.197 $\pm$ 0.001 \\
AtlasNet  \cite{groueix2018atlasnet} &  0.654 $\pm$ 0.001    &  0.197 $\pm$ 0.002\\
FoldingNet \cite{yang2018foldingnet} & 0.666 $\pm$ 0.001  & 0.141 $\pm$  0.001 \\
Ours (PointNet) &  \textbf{0.677 $\pm$ 0.002}   &    \textbf{0.496 $\pm$ 0.004}  \\
\hline
\end{tabular}
\end{center}
\caption{\textbf{Clustering on Aligned and Unaligned Dataset. }}
\label{tab:clusterALigned}
\end{table}

As  seen in  Table \ref{tab:clusterALigned},  our method produces more clusterable embeddings than autoencoder methods. This can be surprising as we are doing instance discrimination and trying to push away every other object in dataset. Our intuition is that, as the neural network is compressing the 3D objects into a lower dimension, the network has to arrange the embeddings strategically which we believe leads to semantic categories being closer in space compared to other objects from different categories.

\subsection{Representation Learning on Rotated Shapes}
In this part, the goal is to test how our method would compare to autoencoder baselines on datasets which are randomly rotated in SO$(3)$ space. As mentioned earlier, most data in real-world scenarios are unaligned. Note that we use rotation in  SO$(3)$ rotation group plus translation as the second view for this section. 

\subsubsection{Simple rotation invariance experiment.}
\begin{figure*}
\centering
\begin{subfigure}[b]{0.3\textwidth}
\includegraphics[trim={1.25cm 1.25cm 1.25cm 1.25cm},clip,width=\textwidth]{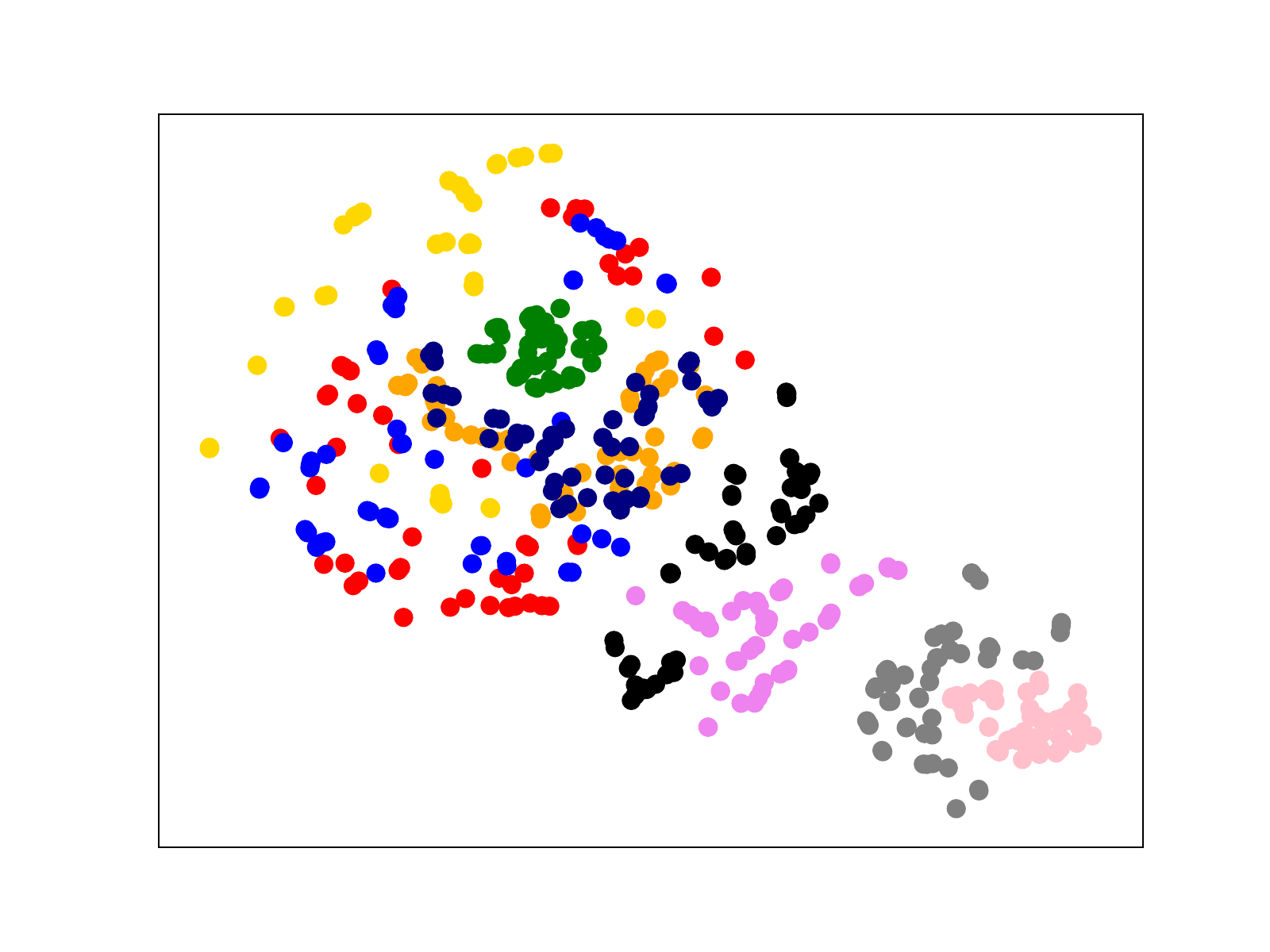}
\caption{Baseline Autoencoder}
\end{subfigure}
\begin{subfigure}[b]{0.3\textwidth}
\includegraphics[trim={1.25cm 1.25cm 1.25cm 1.25cm},clip,width=\textwidth]{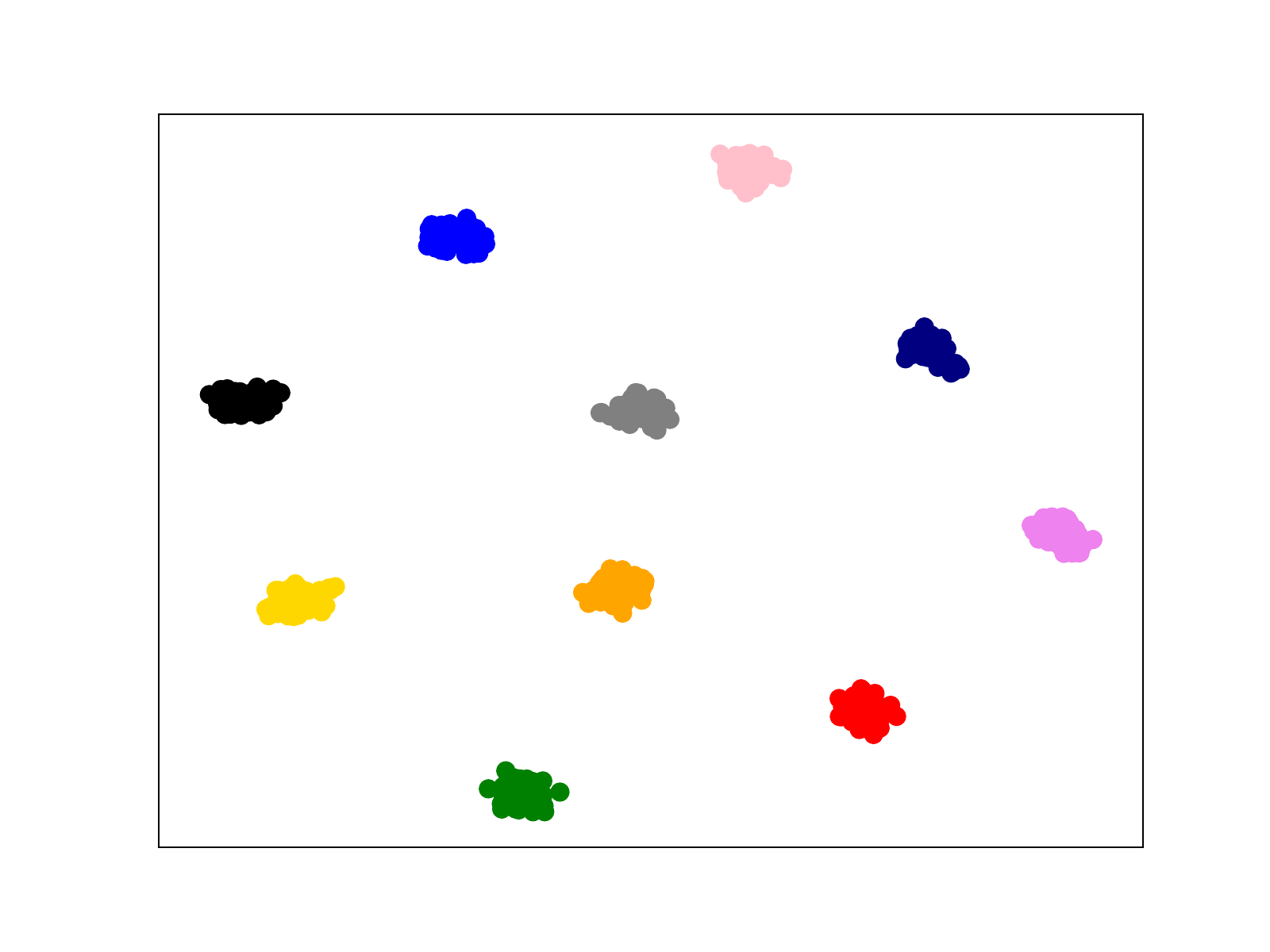}
\caption{Our Model}
\end{subfigure}
\caption{\textbf{t-SNE visualization of rotation invariance check.} Figure illustrating how 3D shapes and their random poses are clustered in our method but fail to cluster in the baseline method.}
\label{fig:tsne_vis}
\end{figure*}

We create a simple experiment to test the sensitivity of the embedding of the shape with respect to the pose of the shape. We conduct the experiment by randomly selecting 10 shapes from ShapeNet v1 dataset and then randomly rotating them 50 times in SO$(3)$ space generating 50 separate objects with different poses per shape. We then generate embeddings for all these objects and then apply clustering. We use t-SNE to give a visual representation of the clustering in $\mathbb{R}^2$ as shown in Figure \ref{fig:tsne_vis}.  We compare with the Latent-GAN \cite{achlioptas2017learning} model.

It can be seen from Figure \ref{fig:tsne_vis} that our model manages to cluster objects and their different poses together. In contrast, Latent-GAN model fails to create meaningful clusters. To quantify this we use the k-means algorithm on the embeddings. In terms of AMI metric, our model achieves 1.0 whereas the baseline achieves 0.555 score. This implies that our method successfully learns to space objects and their different poses in close proximity in latent space, leading to less sensitive embeddings for downstream task.  

\subsubsection{Clusterable representations.} We also test how well our method does on clustering of embeddings when the object is rotated in SO$(3)$ space. The experimental setup is similar to above section and the results are shown in the last column of Table \ref{tab:clusterALigned}. It can be seen that our method significantly out performs the autoencoder baselines. This illustrates that autoencoder baseline are very sensitive to rotation and incorporating some form of rotation invariance into the objective can lead to significant improvement in the embeddings obtained. We also show how this can affect transfer learning results, which are present in the appendix.

\subsubsection{Shape retrieval.} 
\begin{figure*}[!htb]
\begin{center}
\includegraphics[width=0.8\linewidth]{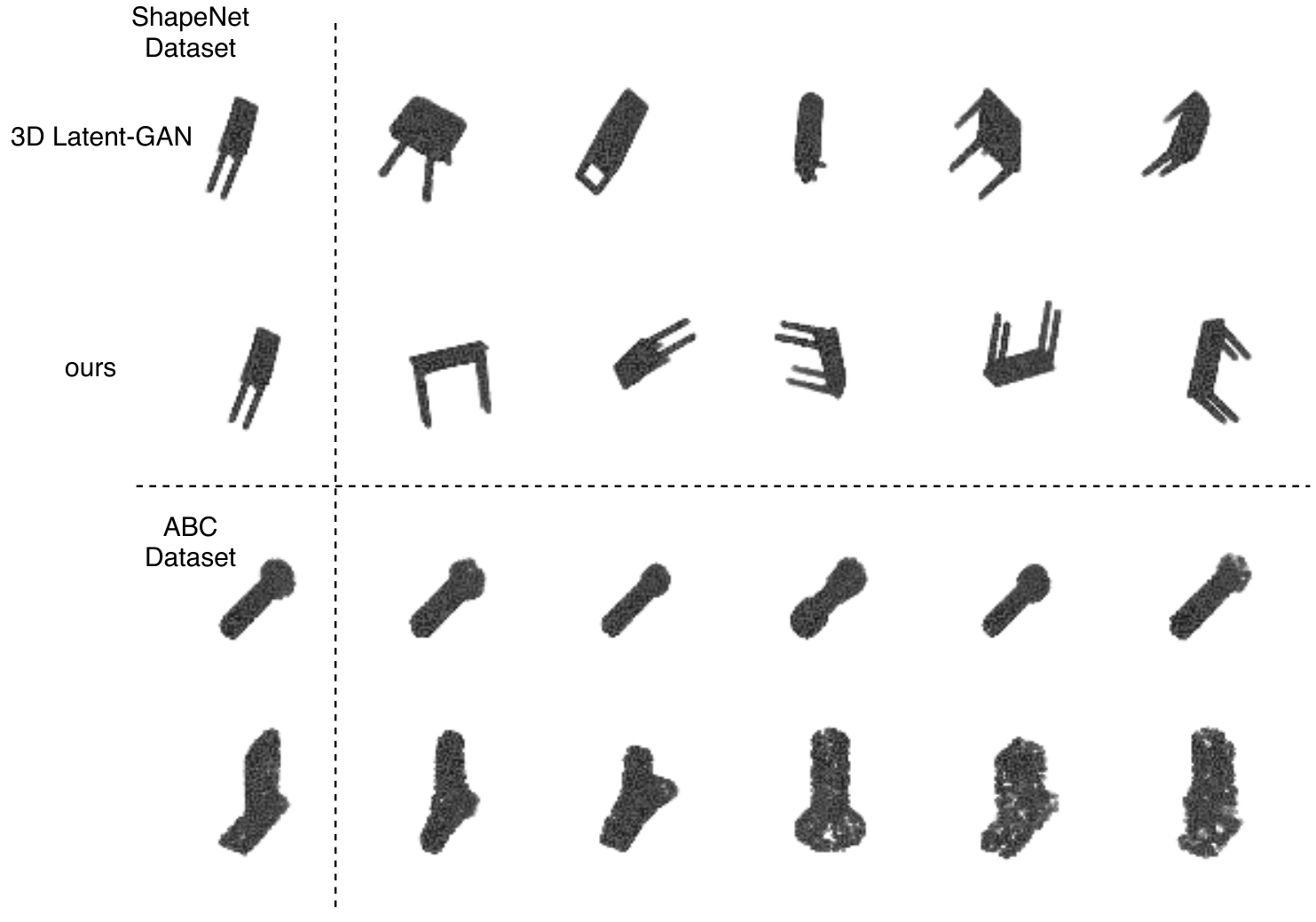}
\end{center}
   \caption{\textbf{Shape Retrieval.} First object in the row represents the query object and next five object are the retrieved objects.  }
\label{fig:main_ret}
\end{figure*}

In many applications, retrieving an object similar to a query object is very useful irrespective of their poses. For such applications, we take  embedding of a query object from ShapeNet v1 dataset and ABC dataset, and retrieve the 5 nearest neighbours for a given shape by using the euclidean distance. The results are shown in Figure \ref{fig:main_ret}. We again compare with Latent-GAN baseline model. 

It can be seen from the first row of Figure \ref{fig:main_ret} that the objects retrieved by the autoencoder baselines are affected by the pose of the object. That is the object retrieved is similar in pose and also sometimes from a different category. Whereas our method manages to bring more semantically similar objects with different poses as shown in second row. The last two rows show sample objects retrieved on ABC dataset. More examples are shown in the appendix.

\subsection{Ablation Studies}

 We detail the affect of using different architectures, transformations and chunks  on the performance of our method. We run most of the experiments on ModelNet40 dataset and use the task of clustering.  We run the clustering algorithm 3 times because of the stochastic nature of k-means. We report the mean and standard deviation on the set of experiments. All the results are measured in AMI.

\subsubsection{Choice of layer}

\begin{table*}[t]
\begin{center}
\tiny
\begin{tabular}{|l|c|c|c|c|c|c|c|}
\hline
Data Type & Layer 1 & Layer 2  & Layer 3 & Layer 4  & Layer 5 & Layer 6 & Layer 7\\
\hline\hline
Aligned  & 0.504 $\pm$ 0.001 & 0.510 $\pm$ 0.002  & 0.517 $\pm$ 0.002 & 0.535 $\pm$ 0.002  & 0.563  $\pm$ 0.004 & \textbf{0.633 $\pm$ 0.002}  & 0.401 $\pm$ 0.005 \\
Unaligned & 0.107 $\pm$ 0.003 & 0.106 $\pm$ 0.002  & 0.112 $\pm$ 0.004 & 0.144 $\pm$ 0.002  & 0.179 $\pm$ 0.002 &  0.404 $\pm$ 0.002   & \textbf{0.496 $\pm$ 0.004} \\
\hline
\end{tabular}
\end{center}
\caption{\textbf{Layer wise embeddings clustering accuracy.}}
\label{tab:layerTable}
\end{table*}

\begin{table}
\begin{center}
\small
\begin{tabular}{|l|c|}
\hline
Data Transformation & Aligned (AMI)  \\
\hline\hline
Translate (T) & 0.633 $\pm$ 0.002 \\

Axis chopped chunk & 0.660 $\pm$ 0.002  \\
Euclidean distance chunk & 0.670 $\pm$ 0.003  \\
Cosine distance chunk & \textbf{0.677 $\pm$ 0.002}   \\
Chebyshev distance chunk & 0.671 $\pm$ 0.001   \\

\hline
\end{tabular}
\end{center}
\caption{\textbf{Effect of different types of augmentation on aligned data of ModelNet40.}}
\label{tab:GeoTransTable}
\end{table}

We see the effect of choosing different layers to obtain the embeddings on the task of clustering. The goal of this experiment is to empirically test which layer contains the most information about the shape. The last layer (7th) of the architecture only consist of a linear layer. We experiment on the aligned as well the unaligned version of ModelNet40 dataset.  The results are shown in Table \ref{tab:layerTable}. 

Based on the results of Table \ref{tab:layerTable}, there are two interesting observations. First, models trained on aligned datasets produce qualitative embeddings from  Layer 6 whereas models on unaligned datasets have informative embedding from layer 7. Hence, we take those respective embeddings for our experiments in the above sections. Secondly, in the case of models trained on unaligned datasets,  Layer 1-5 contain very less clusterable information indicating that using exact position information in pointcloud might not be ideal.

\begin{table}
\begin{center}
\small
\begin{tabular}{|l|c|}
\hline
Data Augmentation  & Unaligned (AMI)  \\
\hline\hline
Rotate SO(3) + Translate  & 0.496 $\pm$ 0.004 \\
Rotate SO(3) + Uniform Scale& 0.313 $\pm$ 0.005 \\
Rotate SO(3) + Random Scale   & 0.407 $\pm$ 0.005 \\
Rotate SO(3) + Uniform Scale + translate & \textbf{0.500 $\pm$ 0.001}\\
Rotate SO(3) + Random Scale  + translate   & 0.485 $\pm$ 0.004 \\

\hline
\end{tabular}
\end{center}
\caption{\textbf{Effect of different types of augmentation on unaligned data of ModelNet40.}}
\label{tab:GeoTransTable2}
\end{table}

\subsubsection{Different types of geometric transformation and chunk selection}
% \subsubsection{Choice of chunk selection}
In this section, we investigate effectiveness of using different ways of obtaining the chunk from a 3D object and geometric transformation of a 3D object for mutual information maximization. We do separate experiments for aligned and unaligned dataset on ModelNet40 dataset.  The transformations for aligned dataset are shown in Table \ref{tab:GeoTransTable} whereas for unaligned dataset it is shown in Table \ref{tab:GeoTransTable2}. In the case of uniform scaling, we scale the object uniformly across the three axis in the range of 0.5 to 1.5 units of the original object. For the translation data augmentation we randomly translation between $-$0.2 to $+$0.2 along each axis. We also do  comparison on the task of transfer learning for aligned dataset as shown in Table \ref{tab:transfer_comp}. Based on the results from the mentioned tables, we choose our transformations for a given task.

\begin{table}
\begin{center}
\small
\begin{tabular}{|l|c|}
\hline
Encoder (Data Augmentation)  & Transfer Learning (Acc. (\%))  \\
\hline\hline
PointNet (Translate) & 87.8 \\
PointNet (Axis chopped chunk) & 89.8   \\
DGCNN (Axis chopped chunk) & \textbf{91.6}    \\
DGCNN (Euclidean distance based chunk) & 90.9    \\
DGCNN (Cosine distance based chunk) & 90.8    \\
DGCNN (Chebyshev distance based chunk) &  91.3   \\
\hline
\end{tabular}
\end{center}
\caption{\textbf{Effect of different types of augmentation on transfer learning of ModelNet40.}}
\label{tab:transfer_comp}
\end{table}

\subsubsection{Effect of chunk size}

The experiment illustrates the trade off between local and global information. If the chunk size is big more global information will be incorporate and the network might fail to capture locality. If the chunk size is small we will capture finer details of the object but will affect the accuracy due to the contrasting nature of the algorithm. The results are shown in Table \ref{tab:KAFFECT}. 

\begin{table}
\begin{center}
\small
\begin{tabular}{|l|c|}
\hline
Chunk size  & Aligned (AMI) \\
\hline\hline
128  &  0.653 $\pm$ 0.002 \\
256  &0.665  $\pm$  0.003 \\
512   & \textbf{0.677 $\pm$ 0.002} \\
768   & 0.658 $\pm$  0.008 \\
1024   & 0.665 $\pm$ 0.001  \\
\hline
\end{tabular}
\end{center}
\caption{\textbf{Effect of different chunk size.}}
\label{tab:KAFFECT}
\end{table}

\section{Conclusion}
In this paper, we investigated using different views of 3D objects to create effective embeddings which generalizes well to different downstream tasks. We showed how considering local substructure in the objective is very effective while considering rotation as a different view can create rotation invariant embeddings. In terms of future work, we would like to explore how  our method generalizes to other tasks such as segmentation and part detection. Secondly, we would like to investigate other views of 3D object such as surface normals. Finally, we would like to extend this method to other 3D representations such as meshes and voxels.

\clearpage
\bibliographystyle{splncs04}
\bibliography{egbib}

\begin{thebibliography}{10}
\providecommand{\url}[1]{\texttt{#1}}
\providecommand{\urlprefix}{URL }
\providecommand{\doi}[1]{https://doi.org/#1}

\bibitem{achlioptas2017learning}
Achlioptas, P., Diamanti, O., Mitliagkas, I., Guibas, L.: Learning
  representations and generative models for 3d point clouds. arXiv preprint
  arXiv:1707.02392  (2017)

\bibitem{anand2019unsupervised}
Anand, A., Racah, E., Ozair, S., Bengio, Y., C{\^o}t{\'e}, M.A., Hjelm, R.D.:
  Unsupervised state representation learning in atari. arXiv preprint
  arXiv:1906.08226  (2019)

\bibitem{bachman2019learning}
Bachman, P., Hjelm, R.D., Buchwalter, W.: Learning representations by
  maximizing mutual information across views. arXiv preprint arXiv:1906.00910
  (2019)

\bibitem{becker1992information}
Becker, S.: An information-theoretic unsupervised learning algorithm for neural
  networks. University of Toronto (1992)

\bibitem{becker1996mutual}
Becker, S.: Mutual information maximization: models of cortical
  self-organization. Network: Computation in neural systems  \textbf{7}(1),
  7--31 (1996)

\bibitem{bell1995information}
Bell, A.J., Sejnowski, T.J.: An information-maximization approach to blind
  separation and blind deconvolution. Neural computation  \textbf{7}(6),
  1129--1159 (1995)

\bibitem{bengio2013representation}
Bengio, Y., Courville, A., Vincent, P.: Representation learning: A review and
  new perspectives. IEEE transactions on pattern analysis and machine
  intelligence  \textbf{35}(8),  1798--1828 (2013)

\bibitem{chang2015shapenet}
Chang, A.X., Funkhouser, T., Guibas, L., Hanrahan, P., Huang, Q., Li, Z.,
  Savarese, S., Savva, M., Song, S., Su, H., et~al.: Shapenet: An
  information-rich 3d model repository. arXiv preprint arXiv:1512.03012  (2015)

\bibitem{chen2020simple}
Chen, T., Kornblith, S., Norouzi, M., Hinton, G.: A simple framework for
  contrastive learning of visual representations. arXiv preprint
  arXiv:2002.05709  (2020)

\bibitem{chen2019learning}
Chen, Z., Zhang, H.: Learning implicit fields for generative shape modeling.
  In: Proceedings of the IEEE Conference on Computer Vision and Pattern
  Recognition. pp. 5939--5948 (2019)

\bibitem{cheng2019meshgan}
Cheng, S., Bronstein, M., Zhou, Y., Kotsia, I., Pantic, M., Zafeiriou, S.:
  Meshgan: Non-linear 3d morphable models of faces. arXiv preprint
  arXiv:1903.10384  (2019)

\bibitem{cohen2016group}
Cohen, T., Welling, M.: Group equivariant convolutional networks. In:
  International conference on machine learning. pp. 2990--2999 (2016)

\bibitem{deng2018ppf}
Deng, H., Birdal, T., Ilic, S.: Ppf-foldnet: Unsupervised learning of rotation
  invariant 3d local descriptors. In: Proceedings of the European Conference on
  Computer Vision (ECCV). pp. 602--618 (2018)

\bibitem{deng20193d}
Deng, H., Birdal, T., Ilic, S.: 3d local features for direct pairwise
  registration. arXiv preprint arXiv:1904.04281  (2019)

\bibitem{esteves2019equivariant}
Esteves, C., Xu, Y., Allen-Blanchette, C., Daniilidis, K.: Equivariant
  multi-view networks. arXiv preprint arXiv:1904.00993  (2019)

\bibitem{groueix2018atlasnet}
Groueix, T., Fisher, M., Kim, V.G., Russell, B.C., Aubry, M.: Atlasnet: A
  papier-m$\backslash$\^{} ach$\backslash$'e approach to learning 3d surface
  generation. arXiv preprint arXiv:1802.05384  (2018)

\bibitem{gutmann2010noise}
Gutmann, M., Hyv{\"a}rinen, A.: Noise-contrastive estimation: A new estimation
  principle for unnormalized statistical models. In: Proceedings of the
  Thirteenth International Conference on Artificial Intelligence and
  Statistics. pp. 297--304 (2010)

\bibitem{hassani2019unsupervised}
Hassani, K., Haley, M.: Unsupervised multi-task feature learning on point
  clouds. In: Proceedings of the IEEE International Conference on Computer
  Vision. pp. 8160--8171 (2019)

\bibitem{henaff2019data}
H{\'e}naff, O.J., Razavi, A., Doersch, C., Eslami, S., Oord, A.v.d.:
  Data-efficient image recognition with contrastive predictive coding. arXiv
  preprint arXiv:1905.09272  (2019)

\bibitem{kingma2014adam}
Kingma, D.P., Ba, J.: Adam: A method for stochastic optimization. arXiv
  preprint arXiv:1412.6980  (2014)

\bibitem{koch2019abc}
Koch, S., Matveev, A., Jiang, Z., Williams, F., Artemov, A., Burnaev, E.,
  Alexa, M., Zorin, D., Panozzo, D.: Abc: A big cad model dataset for geometric
  deep learning. In: Proceedings of the IEEE Conference on Computer Vision and
  Pattern Recognition. pp. 9601--9611 (2019)

\bibitem{lazebnik2004semi}
Lazebnik, S., Schmid, C., Ponce, J.: Semi-local affine parts for object
  recognition (2004)

\bibitem{li2019discrete}
Li, J., Bi, Y., Lee, G.H.: Discrete rotation equivariance for point cloud
  recognition. arXiv preprint arXiv:1904.00319  (2019)

\bibitem{linsker1989application}
Linsker, R.: An application of the principle of maximum information
  preservation to linear systems. In: Advances in neural information processing
  systems. pp. 186--194 (1989)

\bibitem{liu2019relation}
Liu, Y., Fan, B., Xiang, S., Pan, C.: Relation-shape convolutional neural
  network for point cloud analysis. In: Proceedings of the IEEE Conference on
  Computer Vision and Pattern Recognition. pp. 8895--8904 (2019)

\bibitem{mescheder2019occupancy}
Mescheder, L., Oechsle, M., Niemeyer, M., Nowozin, S., Geiger, A.: Occupancy
  networks: Learning 3d struction in function space. In: Proceedings of the
  IEEE Conference on Computer Vision and Pattern Recognition. pp. 4460--4470
  (2019)

\bibitem{michalkiewicz2019deep}
Michalkiewicz, M., Pontes, J.K., Jack, D., Baktashmotlagh, M., Eriksson, A.:
  Deep level sets: Implicit surface representations for 3d shape inference.
  arXiv preprint arXiv:1901.06802  (2019)

\bibitem{misra2019self}
Misra, I., van~der Maaten, L.: Self-supervised learning of pretext-invariant
  representations. arXiv preprint arXiv:1912.01991  (2019)

\bibitem{oord2018representation}
Oord, A.v.d., Li, Y., Vinyals, O.: Representation learning with contrastive
  predictive coding. arXiv preprint arXiv:1807.03748  (2018)

\bibitem{park2019deepsdf}
Park, J.J., Florence, P., Straub, J., Newcombe, R., Lovegrove, S.: Deepsdf:
  Learning continuous signed distance functions for shape representation. arXiv
  preprint arXiv:1901.05103  (2019)

\bibitem{pedregosa2011scikit}
Pedregosa, F., Varoquaux, G., Gramfort, A., Michel, V., Thirion, B., Grisel,
  O., Blondel, M., Prettenhofer, P., Weiss, R., Dubourg, V., et~al.:
  Scikit-learn: Machine learning in python. Journal of machine learning
  research  \textbf{12}(Oct),  2825--2830 (2011)

\bibitem{qi2017pointnet}
Qi, C.R., Su, H., Mo, K., Guibas, L.J.: Pointnet: Deep learning on point sets
  for 3d classification and segmentation. In: Proceedings of the IEEE
  Conference on Computer Vision and Pattern Recognition. pp. 652--660 (2017)

\bibitem{qi2017pointnet++}
Qi, C.R., Yi, L., Su, H., Guibas, L.J.: Pointnet++: Deep hierarchical feature
  learning on point sets in a metric space. In: Advances in neural information
  processing systems. pp. 5099--5108 (2017)

\bibitem{rublee2011orb}
Rublee, E., Rabaud, V., Konolige, K., Bradski, G.R.: Orb: An efficient
  alternative to sift or surf. In: ICCV. vol.~11, p.~2. Citeseer (2011)

\bibitem{sanghitowards}
Sanghi, A., Danielyan, A.: Towards 3d rotation invariant embeddings

\bibitem{sauder2019context}
Sauder, J., Sievers, B.: Context prediction for unsupervised deep learning on
  point clouds. arXiv preprint arXiv:1901.08396  (2019)

\bibitem{steder2010narf}
Steder, B., Rusu, R.B., Konolige, K., Burgard, W.: Narf: 3d range image
  features for object recognition. In: Workshop on Defining and Solving
  Realistic Perception Problems in Personal Robotics at the IEEE/RSJ Int. Conf.
  on Intelligent Robots and Systems (IROS). vol.~44 (2010)

\bibitem{su2015multi}
Su, H., Maji, S., Kalogerakis, E., Learned-Miller, E.: Multi-view convolutional
  neural networks for 3d shape recognition. In: Proceedings of the IEEE
  international conference on computer vision. pp. 945--953 (2015)

\bibitem{sun2019infograph}
Sun, F.Y., Hoffmann, J., Tang, J.: Infograph: Unsupervised and semi-supervised
  graph-level representation learning via mutual information maximization.
  arXiv preprint arXiv:1908.01000  (2019)

\bibitem{tan2018variational}
Tan, Q., Gao, L., Lai, Y.K., Xia, S.: Variational autoencoders for deforming 3d
  mesh models. In: Proceedings of the IEEE Conference on Computer Vision and
  Pattern Recognition. pp. 5841--5850 (2018)

\bibitem{thomas2018tensor}
Thomas, N., Smidt, T., Kearnes, S., Yang, L., Li, L., Kohlhoff, K., Riley, P.:
  Tensor field networks: Rotation-and translation-equivariant neural networks
  for 3d point clouds. arXiv preprint arXiv:1802.08219  (2018)

\bibitem{tian2019contrastive}
Tian, Y., Krishnan, D., Isola, P.: Contrastive multiview coding. arXiv preprint
  arXiv:1906.05849  (2019)

\bibitem{velivckovic2018deep}
Veli{\v{c}}kovi{\'c}, P., Fedus, W., Hamilton, W.L., Li{\`o}, P., Bengio, Y.,
  Hjelm, R.D.: Deep graph infomax. arXiv preprint arXiv:1809.10341  (2018)

\bibitem{wang2019dynamic}
Wang, Y., Sun, Y., Liu, Z., Sarma, S.E., Bronstein, M.M., Solomon, J.M.:
  Dynamic graph cnn for learning on point clouds. ACM Transactions on Graphics
  (TOG)  \textbf{38}(5), ~146 (2019)

\bibitem{worrall2017harmonic}
Worrall, D.E., Garbin, S.J., Turmukhambetov, D., Brostow, G.J.: Harmonic
  networks: Deep translation and rotation equivariance. In: Proceedings of the
  IEEE Conference on Computer Vision and Pattern Recognition. pp. 5028--5037
  (2017)

\bibitem{wu2016learning}
Wu, J., Zhang, C., Xue, T., Freeman, B., Tenenbaum, J.: Learning a
  probabilistic latent space of object shapes via 3d generative-adversarial
  modeling. In: Advances in neural information processing systems. pp. 82--90
  (2016)

\bibitem{wu20153d}
Wu, Z., Song, S., Khosla, A., Yu, F., Zhang, L., Tang, X., Xiao, J.: 3d
  shapenets: A deep representation for volumetric shapes. In: Proceedings of
  the IEEE conference on computer vision and pattern recognition. pp.
  1912--1920 (2015)

\bibitem{wu2018unsupervised}
Wu, Z., Xiong, Y., Yu, S., Lin, D.: Unsupervised feature learning via
  non-parametric instance-level discrimination. arXiv preprint arXiv:1805.01978
   (2018)

\bibitem{yang2018foldingnet}
Yang, Y., Feng, C., Shen, Y., Tian, D.: Foldingnet: Point cloud auto-encoder
  via deep grid deformation. In: Proceedings of the IEEE Conference on Computer
  Vision and Pattern Recognition. pp. 206--215 (2018)

\bibitem{zaheer2017deep}
Zaheer, M., Kottur, S., Ravanbakhsh, S., Poczos, B., Salakhutdinov, R.R.,
  Smola, A.J.: Deep sets. In: Advances in neural information processing
  systems. pp. 3391--3401 (2017)

\bibitem{zhang2019unsupervised}
Zhang, L., Zhu, Z.: Unsupervised feature learning for point cloud understanding
  by contrasting and clustering using graph convolutional neural networks. In:
  2019 International Conference on 3D Vision (3DV). pp. 395--404. IEEE (2019)

\bibitem{zhao20193d}
Zhao, Y., Birdal, T., Deng, H., Tombari, F.: 3d point capsule networks. In:
  Proceedings of the IEEE Conference on Computer Vision and Pattern
  Recognition. pp. 1009--1018 (2019)

\bibitem{zhuang2019local}
Zhuang, C., Zhai, A.L., Yamins, D.: Local aggregation for unsupervised learning
  of visual embeddings. In: Proceedings of the IEEE International Conference on
  Computer Vision. pp. 6002--6012 (2019)

\end{thebibliography}

\end{document}